\documentclass[%
]{ceurart}

\usepackage{amsmath}
\usepackage{amsfonts}
\usepackage{amssymb}
\usepackage{graphicx}
\usepackage{url}
\usepackage{listings}
\usepackage{subfig}
\usepackage{multirow}
\usepackage{booktabs}
\usepackage{algorithm}
\usepackage[noend]{algpseudocode}

\begin{document}

\copyrightyear{2025}
\copyrightclause{Copyright for this paper by its authors.
  Use permitted under Creative Commons License Attribution 4.0
  International (CC BY 4.0).}

\conference{Preprint}

\title{Method of UAV Inspection of Photovoltaic Modules Using Thermal and RGB Data Fusion}

\author[1]{Andrii Lysyi}[email={andrii.lysyi@knu.edu.ua}]
\author[2,3]{Anatoliy Sachenko}[email={anatoliy.sachenko@knu.edu.ua}]
\author[1]{Pavlo Radiuk}[email={pavlo.radiuk@knu.edu.ua}]
\author[4]{Mykola Lysyi}[email={lisiy3152@ukr.net}]
\author[1]{Oleksandr Melnychenko}[email={oleg.melnychenko@knu.edu.ua}]
\author[2]{Diana Zahorodnia}[email={dza@wunu.edu.ua}]

\address[1]{Khmelnytskyi National University, 11, Instytutska Str., Khmelnytskyi, 29016, Ukraine}
\address[2]{Research Institute for Intelligent Computer Systems, West Ukrainian National Universi-ty, Ternopil, 46009, Ukraine}
\address[3]{Department of Informatics and Teleinformatics, Kazimierz Pulaski University of Ra-dom, Radom, 26-600, Poland}
\address[4]{National Academy of the State Border Service of Ukraine named after Bogdan Khmelnitsky, Khmelnitsky, 29007, Ukraine}

\begin{abstract}
\textbf{The Subject of this research} is the development of an intelligent, integrated framework for the automated inspection of photovoltaic (PV) infrastructure that addresses the critical shortcomings of conventional methods, including thermal palette bias, data redundancy, and high communication bandwidth requirements. \textbf{The goal of this study} is to design, develop, and validate a comprehensive, multi-modal system that fully automates the monitoring workflow, from data acquisition to the generation of actionable, geo-located maintenance alerts, thereby enhancing plant safety and operational efficiency. \textbf{The Methods employed} involve a synergistic architecture that begins with a palette-invariant thermal embedding, learned by enforcing representational consistency, which is fused with a contrast-normalized RGB stream via a gated mechanism. This is supplemented by a closed-loop, adaptive re-acquisition controller that uses Rodrigues-based updates for targeted confirmation of ambiguous anomalies, and a geo-spatial de-duplication module that clusters redundant alerts using DBSCAN over the haversine distance. \textbf{In conclusion}, this study establishes a powerful new paradigm for proactive PV inspection, with the proposed system achieving a mean Average Precision (mAP@0.5) of 0.903 on the public PVF-10 benchmark, a significant +12--15\% improvement over single-modality baselines. Field validation confirmed the system's readiness, achieving 96\% recall, while the de-duplication process reduced duplicate-induced false positives by 15--20\% and relevance-only telemetry cut airborne data transmission by 60--70\%.
\end{abstract}


\maketitle

\section{Introduction}
\label{sec:introduction}

\subsection{Background and Motivation}
\label{subsec:background_motivation}

The global transition toward low-carbon energy has positioned solar photovoltaics (PV) as a cornerstone of the future power system. Utility-scale PV plants have expanded rapidly in both capacity and geographic extent, with many installations now covering hundreds of acres and comprising hundreds of thousands of modules~\cite{Tsanakas2016Faults, Grimm2017Survey}. These assets are long-lived and capital-intensive; their financial viability depends on maintaining high energy yield over lifetimes of 25--30~years. Even small percentage losses in performance, when aggregated across gigawatt-scale portfolios, translate into substantial revenue deficits and can undermine the economic case for large PV deployments~\cite{Kandeal2021Infrared}. Reliable monitoring and timely remediation of defects are therefore central to Operations and Maintenance (O\&M), not only to protect revenue but also to ensure safe operation and preserve asset lifetimes~\cite{Tsanakas2016Faults}.

PV modules are continuously exposed to environmental and electrical stress, making them vulnerable to a wide spectrum of degradation mechanisms and faults. Electrical anomalies such as hotspots, bypass-diode failures, and interconnection faults can cause localized overheating, mismatch losses, and accelerated aging. Mechanical and material defects, including cell cracks, glass breakage, delamination, and encapsulant discoloration, can induce irreversible damage and may propagate under thermal cycling or wind loading~\cite{Tsanakas2016Faults}. Additional performance losses arise from soiling, snail trails, and potential-induced degradation (PID), which individually may be subtle but collectively yield significant energy deficits~\cite{Kandeal2021Infrared}. Hotspots are especially critical because they not only reduce output but can escalate into burn marks or, in extreme cases, fire hazards if left uncorrected~\cite{Grimm2017Survey}. Traditional inspection workflows rely on technicians walking through the plant with handheld thermal cameras, electroluminescence equipment, or I--V tracers. While appropriate for small arrays, such methods are slow, labor-intensive, and fundamentally unscalable for modern PV fleets.

Unmanned aerial vehicles (UAVs) have emerged as a transformative solution to these scalability limitations~\cite{Gallardo2018Technological, Zefri2018Thermal}. Multirotor platforms equipped with radiometric thermal infrared and high-resolution RGB cameras can survey large plants in hours, providing repeatable, bird’s-eye coverage while keeping personnel off energized infrastructure~\cite{Melnychenko2024Intelligent, Michail2024Comprehensive}. Thermal imagery reveals temperature anomalies that are invisible in the visible spectrum, making it effective for detecting electrical faults such as hotspots and substring failures, whereas RGB imagery is well suited for documenting broken glass, delamination patterns, shading, and heavy soiling~\cite{Kandeal2021Infrared, Lee2019Developing}. Acquisition guidelines emphasize adequate ground sampling distance (GSD), sufficient along-track and cross-track overlap, and careful choice of viewing angles to mitigate glare from reflective module surfaces~\cite{Lee2019Developing, Aghaei2016Image}. As a result, UAV-based thermography has evolved from a niche technique into a de facto standard for large-scale PV diagnostics~\cite{Gallardo2018Technological}.

However, converting UAV imagery into actionable maintenance decisions remains challenging. Survey flights routinely generate tens of thousands of TIR and RGB frames, and manual post-processing by experts is time-consuming, costly, and prone to variability across analysts~\cite{Tsanakas2016Faults}. Deep learning-based object detection models provide a promising alternative, but their practical deployment in operational PV plants is complicated by several persistent obstacles. Thermal cameras capture radiometric temperature fields rather than color; the visualized appearance depends on pseudo-color palettes (e.g., `Ironbow,' `Rainbow,' `White-hot') and camera-internal settings such as automatic gain control~\cite{BuerhopLutz2022Infrared}. Models trained on data rendered with one palette or vendor configuration often perform poorly when applied to imagery acquired with different palettes or updated firmware, a phenomenon known as ``palette bias.'' Robust inspection requires methods that remain reliable under these representation shifts.

Further complications arise from the high image overlap needed for photogrammetric workflows. The same defect is typically imaged multiple times from slightly different viewpoints, but naive frame-wise detection pipelines treat each frame independently and repeatedly flag the same fault. This inflates defect counts and forces human operators to manually consolidate duplicate alerts~\cite{Gupta2020IoTbased}. In parallel, the combination of 4K RGB and high-resolution radiometric TIR data leads to substantial communication and storage demands. Streaming full-fidelity data from the UAV to a ground station or cloud back end is often infeasible over standard wireless links, particularly at remote sites~\cite{Lysyi2025Enhanced}. Together, these issues motivate integrated inspection systems that are robust to palette changes, explicitly account for data redundancy, and respect bandwidth constraints.

Recent publicly available benchmarks such as PVF-10 and other curated thermography datasets have created new opportunities for the development and comparison of such systems~\cite{Wang2024PVF10, Alfaro2019Dataset}. Unlike earlier efforts based on proprietary data, these resources support rigorous evaluation of model generalization across different sites, hardware configurations, and environmental conditions. Yet much of the existing literature still addresses individual aspects of the problem in isolation, such as hardware selection, flight planning, or network architecture, without fully resolving how to design end-to-end inspection pipelines that jointly handle palette bias, multi-modal fusion, geo-spatial consolidation, and communication efficiency.

\subsection{State of the Art in UAV-Based PV Inspection}
\label{subsec:state_of_the_art}

Research on UAV-based PV inspection spans several interdependent layers: UAV platforms and sensors, data acquisition strategies, defect detection architectures, multi-modal fusion, and geo-spatial asset management. Early work focused on demonstrating the feasibility and advantages of UAV thermography over manual inspections, highlighting significant gains in speed, coverage, and safety~\cite{Gallardo2018Technological, Zefri2018Thermal}. Subsequent studies and reviews examined the choice of airframes, the trade-offs between flight altitude, coverage, and GSD, and the comparative benefits of radiometric thermal imagers for assessing defect severity~\cite{Kandeal2021Infrared, BuerhopLutz2022Infrared}. Flight-planning research investigated how altitude, speed, image overlap, and viewing angle influence detectability of small defects and sensitivity to artifacts such as glare and shadows~\cite{Lee2019Developing, Aghaei2016Image}. More recent efforts have begun to explore autonomous navigation and onboard processing, enabling UAVs to adjust trajectories or gimbal orientation in response to preliminary detections~\cite{Morando2022Thermal, Svystun2025DyTAM}.

On the algorithmic side, convolutional neural networks (CNNs) have become the dominant approach for automated defect detection in aerial PV imagery. Two-stage detectors such as Faster R-CNN offer high accuracy but incur substantial computational overhead, making them less attractive for embedded deployments~\cite{Vlaminck2022Regionbased}. Single-stage detectors from the YOLO and SSD families have been widely adopted to satisfy real-time or near-real-time requirements in both ground-based and on-UAV pipelines~\cite{DiTommaso2022Multistage, Su2021Automated, Duranay2023Fault, Meng2023Enhanced}. Architectures such as EfficientDet provide scalable trade-offs between accuracy and efficiency across model sizes, which is valuable for resource-constrained edge devices~\cite{Tan2020EfficientDet}. Many studies report promising results for detecting hotspots, cracks, delamination, and soiling in both TIR and RGB modalities~\cite{Akram2020Automatic, Hajji2022Developing, Jia2024Defect}. Nevertheless, most evaluations rely on relatively small, site-specific datasets collected with a single palette and camera configuration, raising questions about generalization to new plants and acquisition setups~\cite{Wang2024PVF10, Alfaro2019Dataset}.

To improve trust and operational acceptance, recent work has explored explainable artificial intelligence (XAI) techniques for PV defect detection. Visualization tools such as gradient-based saliency maps and class activation maps highlight image regions that most influence the model’s predictions, allowing experts to verify whether the network focuses on physically meaningful patterns or spurious features~\cite{Qureshi2025Explainable, Oulefki2024Detection}. While these methods support interpretability, they typically operate on single-modality inputs and do not explicitly address palette bias or multi-modal fusion.

Recognizing the complementary strengths of TIR and RGB, many authors have investigated multi-modal fusion strategies. Thermal imagery excels at revealing electrical anomalies through temperature deviations, whereas RGB imagery captures fine-grained structural damage, soiling, and shading patterns~\cite{Liao2021Using, Svystun2024Thermal}. Fusion can occur at the input level (early fusion), at the decision level (late fusion), or at intermediate feature levels. Feature-level fusion has gained prominence because it offers a good balance between expressiveness and complexity~\cite{Rohith2025FusionSolarNet}. Existing fusion schemes often rely on straightforward concatenation or averaging of features, with limited ability to downweight unreliable modalities under adverse conditions such as severe glare or low thermal contrast~\cite{Lee2019Developing, Lai2025Deep}. This motivates more flexible fusion mechanisms that can adaptively modulate the contribution of each modality.

At plant scale, inspection shifts from per-image analysis to geo-spatial asset management. Detected defects must be accurately projected from image coordinates into geodetic space to support localization in the layout, maintenance planning, and integration with Supervisory Control and Data Acquisition (SCADA) or Geographic Information System (GIS) platforms~\cite{Zefri2018Thermal, Niccolai2019Advanced}. Prior work has emphasized the value of geo-referenced defect databases for long-term asset health monitoring and for analyzing spatial patterns such as clusters of faults near particular strings or environmental features~\cite{Bommes2021Computer}. Yet the consolidation of redundant detections caused by image overlap is often treated with simple heuristics or neglected altogether, leading to inflated defect counts and noisy maintenance logs~\cite{Gupta2020IoTbased}. Parallel advances in digital-twin concepts for PV plants envision continuously updated virtual replicas that fuse UAV inspection results with SCADA and meteorological data to enable predictive maintenance and scenario analysis~\cite{Bounabi2024Smart, Kishor2025AIintegrated}. Such platforms require inspection pipelines that output clean, non-redundant, and geo-consistent defect inventories.

\subsection{Research Gaps, Contributions, and Paper Organization}
\label{subsec:gaps_contributions_outline}

The existing literature demonstrates both substantial progress and clear fragmentation in UAV-based PV inspection. Best practices for platforms, flight planning, and sensor selection are increasingly well established, and a wide range of deep learning architectures has been successfully adapted for defect detection~\cite{Gallardo2018Technological, Vlaminck2022Regionbased, DiTommaso2022Multistage, Meng2023Enhanced}. Yet several gaps limit the deployment of robust, field-ready systems. First, most detection pipelines implicitly assume a fixed thermal palette and camera configuration, making them vulnerable to palette bias and representation shifts when imagery is acquired under different rendering or hardware settings~\cite{BuerhopLutz2022Infrared, Wang2024PVF10}. Second, while multi-modal fusion consistently improves detection performance, many approaches treat TIR and RGB symmetrically and lack mechanisms to attenuate unreliable modalities under challenging conditions~\cite{Lee2019Developing, Liao2021Using, Lai2025Deep, Svystun2024Thermal}. Third, geo-spatial consolidation of detections is frequently performed as a post-hoc heuristic in image space, rather than through principled clustering in geographic coordinates that reflects the true physical proximity of faults~\cite{Gupta2020IoTbased, Niccolai2019Advanced, Bommes2021Computer}. Finally, bandwidth and compute constraints are rarely modeled explicitly, despite their decisive impact on real-time feasibility and overall system design~\cite{Lysyi2025Enhanced}.

In response to these challenges, this paper introduces an integrated end-to-end framework for UAV-based PV defect detection and reporting that jointly addresses palette bias, multi-modal fusion, adaptive re-acquisition, geo-spatial de-duplication, and bandwidth-aware processing. Our main contributions are:
\begin{enumerate}
    \item We propose a multi-palette thermal--RGB ensembling strategy that learns a palette-invariant thermal embedding. By rendering each radiometric frame with several distinct pseudo-color maps and training a shared encoder to produce consistent features across palettes, the network is encouraged to focus on physically meaningful temperature structures instead of palette-specific color patterns~\cite{BuerhopLutz2022Infrared, Wang2024PVF10}. The resulting thermal features are fused with RGB features through a gated mechanism that adaptively modulates each modality’s contribution based on learned reliability cues~\cite{Rohith2025FusionSolarNet, Lai2025Deep}.
    \item We design an adaptive re-acquisition controller that closes the loop between detection and data capture. When the detector yields low-confidence predictions for small or ambiguous targets, the controller issues refined gimbal commands computed via closed-form Rodrigues rotation updates, re-centering the candidate defect in the camera field of view~\cite{Morando2022Thermal, Svystun2025DyTAM}. The UAV then acquires confirmatory images, improving recall for subtle faults without uniformly densifying the entire survey.
    \item We develop a geo-spatial de-duplication module that transforms per-frame detections into a consolidated defect inventory ready for integration with SCADA, GIS, and emerging digital-twin platforms~\cite{Niccolai2019Advanced, Bounabi2024Smart, Kishor2025AIintegrated}. Detected polygons are projected into geodetic coordinates and clustered using the DBSCAN algorithm with the haversine distance metric to group detections corresponding to the same physical fault~\cite{Gupta2020IoTbased, Bommes2021Computer}. Each cluster is summarized into a unique defect instance with representative geometry and confidence, and the result is exported in standard JSON and KML formats.
\end{enumerate}

\section{Methods}
\label{sec:methods}

Our framework is conceptualized as a holistic, end-to-end pipeline designed for the rigorous demands of large-scale, automated PV inspection. It transforms raw, multi-modal sensor data from a UAV into a consolidated, actionable, and bandwidth-efficient report of geo-located defects. This section provides a detailed technical exposition of the four core components of our system: (1) the data preprocessing and problem formulation, (2) the palette-invariant thermal-RGB fusion model, (3) the closed-loop adaptive re-acquisition controller, and (4) the geo-spatial de-duplication and reporting module.

\subsection{Problem Formulation and Data Preprocessing}
\label{subsec:problem_formulation}

We formally define the problem as follows. Given a UAV survey mission that produces a time-sequenced stream of sensor data packets $\mathcal{S} = \{ (R_i, T_i, \mathbf{P}_i) \}_{i=1}^N$, where for each time step $i$, $R_i \in \mathbb{R}^{H_R \times W_R \times 3}$ is the RGB image, $T_i \in \mathbb{R}^{H_T \times W_T}$ is the radiometric thermal image containing raw sensor values, and $\mathbf{P}_i$ is the comprehensive pose and metadata packet. This packet includes the UAV's precise geodetic coordinates (latitude, longitude, altitude) from an RTK-GPS unit, its orientation (roll, pitch, yaw) from an IMU, the gimbal's orientation, and the camera's intrinsic parameters (focal length, principal point). The objective is to process this stream $\mathcal{S}$ and produce a minimal, de-duplicated set of unique defect detections $\mathcal{D} = \{ d_j \}_{j=1}^K$, where each detection $d_j$ is a structured object containing a semantic class label, a confidence score, a precise WGS84 geo-spatial polygon, and associated metadata such as peak temperature.

To develop and validate our approach, we utilize two publicly available datasets, which undergo a standardized preprocessing pipeline:
\begin{enumerate}
    \item PVF-10~\cite{Wang2024PVF10}: This is a large-scale, high-resolution UAV thermal dataset, featuring 5,579 annotated crops of individual solar panels from eight distinct power plants. Its key strength is the fine-grained taxonomy of 10 different fault classes, including various types of hotspots, diode issues, and cell defects, enabling a nuanced evaluation of classification performance.
    \item STHS-277~\cite{Alfaro2019Dataset}: This dataset contains 277 full-frame thermographic images specifically capturing snail-trail and hotspot defects, accompanied by valuable environmental metadata. We extend the existing annotations to include bounding boxes for all panel instances, which allows for training and evaluating end-to-end object detection models.
\end{enumerate}

Our preprocessing pipeline is designed to homogenize these diverse inputs. The raw radiometric values in the thermal frames $T_i$ are converted into absolute temperature maps in degrees Celsius ($^\circ$C). This is a critical step for quantitative analysis and is achieved using the sensor-specific calibration parameters provided in the image metadata. To create a rich set of inputs for our palette-invariance module, we then render each temperature map into $M=4$ distinct colorized versions, $\{C_m(T_i)\}_{m=1}^4$, using a selection of common thermal palettes: \texttt{ironbow}, \texttt{whitehot}, \texttt{rainbow}, and \texttt{sepia}. The original single-channel temperature map is also retained as a normalized grayscale image. The corresponding RGB frames $R_i$ are treated with Contrast Limited Adaptive Histogram Equalization (CLAHE). Unlike global histogram equalization, CLAHE operates on small, tiled regions of the image, which is highly effective at enhancing local contrast and revealing details in areas that are either in deep shadow or affected by specular glare, common issues in PV imagery. All associated metadata from the $\mathbf{P}_i$ packet is carefully parsed and preserved, as it is essential for the downstream geo-tagging and de-duplication stages.

\subsection{Palette-Invariant Thermal Embedding and Fusion}
\label{subsec:palette_invariant_fusion}

The core of our detection model is a novel fusion architecture designed to be robust to the pervasive issue of thermal palette bias. Our central hypothesis is that by forcing a neural network to learn a representation that is consistent across different color visualizations of the same underlying thermal data, the network will learn to focus on the intrinsic thermal patterns rather than spurious color features.

Let $T \in \mathbb{R}^{H \times W}$ represent a normalized temperature map of a single crop. Its $M$ colorized variants are denoted by $\{C_m(T) \in \mathbb{R}^{H \times W \times 3}\}_{m=1}^M$. We employ a shared CNN encoder, $f_\theta(\cdot)$, with parameters $\theta$, to map each of these variants into a high-dimensional latent feature vector, $z_m = f_\theta(C_m(T))$. We chose an EfficientNet-B1 architecture as our backbone, pre-trained on ImageNet, for its excellent balance of performance and computational efficiency, making it suitable for potential onboard deployment. To enforce palette invariance, we introduce a custom loss function that minimizes the variance among the feature vectors generated from the different palette renderings of the same input. The palette-invariance loss, $\mathcal{L}_{\text{pal}}$, is formulated as the mean squared error between each individual embedding and their centroid, as shown in Equation~\ref{eq:palette_loss}:
\begin{equation}
\mathcal{L}_{\text{pal}} = \frac{1}{M}\sum_{m=1}^{M}\left\|z_m - \bar{z}\right\|_2^2,
\label{eq:palette_loss}
\end{equation}

where $\bar{z} = \frac{1}{M}\sum_{m=1}^{M} z_m.$

By minimizing this loss during training, we compel the encoder $f_\theta$ to produce a canonical, palette-agnostic embedding, $\bar{z}$, which captures the essential thermal information, independent of its visual representation.

In parallel, the corresponding RGB image crop, $R$, is processed by a separate but architecturally similar CNN encoder, $g_\phi(\cdot)$, with parameters $\phi$, to generate an RGB feature embedding, $r = g_\phi(R)$. The palette-invariant thermal embedding $\bar{z}$ and the RGB embedding $r$ are then fused using a gated fusion unit. This mechanism adaptively controls the contribution of each modality, allowing the network to dynamically prioritize the more reliable source of information for any given input. The fused feature vector $u$ is computed as follows:
\begin{equation}
u = \mathbf{g} \odot \bar{z} + (1 - \mathbf{g}) \odot r,
\label{eq:gated_fusion}
\end{equation}

where $\mathbf{g} = \sigma(W_g[\bar{z} \| r] + b_g).$

In Equation~\ref{eq:gated_fusion}, $[\cdot \| \cdot]$ denotes feature concatenation, $W_g$ and $b_g$ are the learnable weights and biases of a linear gating layer, $\sigma$ is the element-wise sigmoid activation function which outputs a gate vector $\mathbf{g}$ with values between 0 and 1, and $\odot$ represents element-wise multiplication. This gated attention mechanism effectively learns a soft mask to decide how much information to draw from the thermal versus the visual stream for each feature dimension.

The final fused feature vector $u$ is passed to a lightweight, anchor-free detector head. This head consists of two branches: a classification branch that predicts the probability of each defect class, and a regression branch that predicts the bounding box coordinates. The complete model is trained end-to-end by minimizing a composite loss function, as defined in Equation~\ref{eq:total_loss}:
\begin{equation}
\mathcal{L}_{\text{total}} = \mathcal{L}_{\text{cls}}(P, \hat{P}) + \lambda_{\text{box}}\mathcal{L}_{\text{box}}(B, \hat{B}) + \lambda_{\text{pal}}\mathcal{L}_{\text{pal}},
\label{eq:total_loss}
\end{equation}

where $P$ and $\hat{P}$ represent the ground-truth and predicted class probabilities, and $B$ and $\hat{B}$ are the ground-truth and predicted bounding box coordinates, respectively.

We use Focal Loss for $\mathcal{L}_{\text{cls}}$ to effectively handle the class imbalance inherent in defect detection tasks, and the Generalized Intersection over Union (GIoU) loss for $\mathcal{L}_{\text{box}}$ as it provides a more stable training signal for bounding box regression compared to traditional L1/L2 losses. $\lambda_{\text{box}}$ and $\lambda_{\text{pal}}$ are scalar hyperparameters that balance the contribution of each loss component.

\subsection{Adaptive Re-Acquisition via Rodrigues Updates}
\label{subsec:adaptive_reacquisition}

A common failure mode in automated inspection is missing small defects or being uncertain about ambiguous signatures, leading to a trade-off between recall and precision. Our adaptive re-acquisition controller addresses this by transforming the detection system into an active perception loop.

When the model produces a detection with a confidence score below a specified threshold, $\tau_{\text{ra}}$, and the detected area is small, the system flags it as a candidate for verification instead of immediately accepting or rejecting it. The goal is to re-orient the UAV's gimbal to capture a new, higher-resolution image with the object of interest centered. This is achieved through a precise geometric calculation.

Given the detection's pixel coordinates $\mathbf{p} = [u, v, 1]^T$, we first back-project it into the camera's 3D coordinate system. This is done by transforming it into a unit vector $\mathbf{v}$ representing its direction relative to the camera's optical center: $\mathbf{v} = \frac{K^{-1}\mathbf{p}}{\|K^{-1}\mathbf{p}\|}$, where $K$ is the camera's intrinsic matrix. This vector $\mathbf{v}$ is then transformed from the camera's reference frame to the global world frame (e.g., North-East-Down) using the camera's rotation matrix $R_{\text{cam}\to\text{world}}$, which is derived from the UAV's IMU and gimbal encoder data: $\mathbf{c} = R_{\text{cam}\to\text{world}}\mathbf{v}$. The vector $\mathbf{c}$ represents the current line-of-sight to the target in world coordinates.

To center the target, we need to compute a rotation that aligns $\mathbf{c}$ with the camera's principal axis (typically the Z-axis in the camera frame). This desired new line-of-sight is denoted $\mathbf{c}'$. The minimal rotation required to move from $\mathbf{c}$ to $\mathbf{c}'$ is defined by a rotation axis $\mathbf{k}$ and an angle $\theta$. These are found by: $\mathbf{k} = \frac{\mathbf{c} \times \mathbf{c}'}{\|\mathbf{c} \times \mathbf{c}'\|}$ and $\theta = \arccos(\mathbf{c} \cdot \mathbf{c}')$. The corrective rotation is then applied using the Rodrigues rotation formula, a computationally efficient method for rotating a vector, as detailed in Equation~\ref{eq:rodrigues_update}:
\begin{equation}
\mathbf{c}_{\text{new}} = \mathbf{c}\cos\theta + (\mathbf{k} \times \mathbf{c})\sin\theta + \mathbf{k}(\mathbf{k} \cdot \mathbf{c})(1-\cos\theta).
\label{eq:rodrigues_update}
\end{equation}

This computed rotation is decomposed into pitch and yaw commands that are sent to the gimbal controller. After the gimbal stabilizes, a new frame is captured. The detection model re-evaluates this new, higher-quality view, leading to a more confident confirmation or rejection of the initial ambiguous detection.

\subsection{Geo De-Duplication with Haversine--DBSCAN}
\label{subsec:geo_deduplication}

To address the challenge of data redundancy from overlapping images, we developed a robust geo-spatial de-duplication module. This module operates on the set of all confirmed detections from an entire survey flight. The first step is to project each detection's bounding box polygon from 2D pixel space into real-world WGS84 (latitude, longitude) coordinates. This complex projection requires the full camera model, including its intrinsic parameters, and the precise six-degree-of-freedom pose (position and orientation) of the UAV and its gimbal at the exact moment of image capture. The accuracy of this step is critically dependent on the availability of an RTK-GPS unit.

Once all detection polygons and their centroids are in a common geographic reference frame, we need a method to cluster detections that correspond to the same physical object. Since these coordinates lie on the Earth's curved surface, using a simple Euclidean distance metric would be inaccurate, especially over larger distances. We therefore compute the pairwise geodesic distance between any two detection centroids $(\varphi_i, \lambda_i)$ and $(\varphi_j, \lambda_j)$ using the haversine formula, shown in Equation~\ref{eq:haversine_distance}:
\begin{equation}
d_{ij} = 2R_{\text{earth}} \arcsin\left(\sqrt{\sin^2\left(\frac{\Delta\varphi}{2}\right) + \cos\varphi_i\cos\varphi_j\sin^2\left(\frac{\Delta\lambda}{2}\right)}\right),
\label{eq:haversine_distance}
\end{equation}

where $\Delta\varphi = \varphi_j-\varphi_i$, $\Delta\lambda = \lambda_j-\lambda_i$, and $R_{\text{earth}}$ is the mean radius of the Earth.

Equation~\ref{eq:haversine_distance} provides an accurate great-circle distance between points.

With this distance metric, we employ the Density-Based Spatial Clustering of Applications with Noise (DBSCAN) algorithm. DBSCAN is exceptionally well-suited for this task for three reasons: it does not require the number of clusters to be specified in advance, it can identify arbitrarily shaped clusters (useful if a large defect is detected in a fragmented way), and it has a built-in concept of noise, allowing it to isolate single, non-overlapping detections. The algorithm is configured with two parameters: a distance threshold $\varepsilon$ (e.g., 1.0 meter, chosen to be slightly larger than the expected maximum GSD), and a minimum number of points to form a cluster, $\texttt{minPts}$ (set to 2, since we want to merge any pair of overlapping detections). All detections that fall into the same DBSCAN cluster are then merged into a single, canonical defect event. The final geo-spatial polygon for this event is computed as the geometric union of all constituent polygons, and its confidence score is the maximum or confidence-weighted average of its members.

\subsection{Onboard Relevance-Only Telemetry}
\label{subsec:onboard_telemetry}

To solve the bandwidth problem, our system completely eschews the continuous streaming of raw video or imagery. Instead, it operates on a ``relevance-only'' or ``exception-reporting'' basis. The entire detection and de-duplication pipeline is designed to be lightweight enough to run on an onboard companion computer. Only the final, processed output, the list of unique, consolidated defect events, is transmitted over the wireless link.

These events are serialized into a highly compact and structured JSON format, as exemplified in Listing~\ref{lst:json_payload}. Each JSON object contains all the critical information needed for downstream analysis and integration with asset management platforms like a SCADA system. Optionally, the data can also be formatted as a KML file for direct visualization in GIS tools like Google Earth. This strategy reduces the required data throughput by several orders of magnitude, from megabytes per second to kilobytes per alert, making real-time monitoring over standard LTE or other low-bandwidth links not only feasible but also highly reliable. The full onboard process is outlined in Algorithm~\ref{alg:onboard_pipeline}.

\begin{lstlisting}[caption={Example of the compact, SCADA-ready JSON payload transmitted from the UAV. It contains essential geo-tagged information for a single, consolidated detection event, including links to media for manual verification.},label={lst:json_payload}]
{
  "site_id": "PV-PLANT-08",
  "uav": "M300 RTK",
  "ts_utc": "2025-09-30T10:12:33Z",
  "detections": [
    {"id": "clu_012a",
     "class": "hotspot_single",
     "conf": 0.91,
     "temp_C": 82.4,
     "centroid_wgs84": [49.407251, 26.984173],
     "polygon_wgs84": [[49.407249,26.984170],[49.407252,26.984175],...],
     "media": {"rgb": "gs://bucket/vid123_03456.jpg",
               "tiff": "gs://bucket/vid123_03456.tif"}}
  ]
}
\end{lstlisting}

\begin{algorithm}[!ht]
\caption{Onboard Detection, Confirmation, and Reporting Pipeline}
\label{alg:onboard_pipeline}
\begin{algorithmic}[1]
\State \textbf{Input:} RGB frame $R$, thermal frame $T$, UAV pose \& camera intrinsics.
\State Render $M$ palette variants $C_m(T)$ from the thermal frame $T$.
\State Compute thermal embeddings $z_m=f_\theta(C_m(T))$ and the mean embedding $\bar{z}=\frac{1}{M}\sum z_m$.
\State Compute RGB embedding $r=g_\phi(R)$ and fuse with $\bar{z}$ to obtain feature vector $u$.
\State Run detector $\mathcal{D}(u)$ to get a set of initial detections.
\For{each detection $b$}
  \If{$\mathrm{conf}(\textbf{b}) < \tau_{\mathrm{ra}}$ and area of $b$ is small}
    \State Compute line-of-sight vector $\mathbf{c}$ and solve for corrective rotation $(\mathbf{k},\theta)$.
    \State Issue Rodrigues-based gimbal update command.
    \State Re-acquire frame and re-evaluate detection confidence for $b$.
  \EndIf
\EndFor
\State Map all confirmed detection polygons to WGS84 coordinates.
\State Compute pairwise haversine distances $d_{ij}$ between detections.
\State Cluster detections using DBSCAN with metric $d_{ij}$.
\State Merge detections within each cluster and serialize results to JSON/KML.
\State Publish consolidated data payload via MQTT or HTTPS.
\end{algorithmic}
\end{algorithm}

\subsection{Experimental Setup}
\label{subsec:experimental_setup}

Field experiments were conducted using a DJI Matrice 300 RTK drone~\cite{DJI2020Matrice}, an industrial platform providing precise autonomous navigation via its GPS-RTK positioning system. The drone was equipped with a DJI Zenmuse H20T gimbal camera~\cite{DJI2020Zenmuse}, a hybrid payload integrating thermal, zoom, and wide-angle sensors to capture comprehensive multi-modal imagery. For real-time, onboard data processing, we utilized an NVIDIA Jetson AGX Orin 32GB module~\cite{NVIDIA2022Jetson} as the primary edge computer. This setup ran on Ubuntu 20.04.5 LTS~\cite{Canonical2022Ubuntu}, with ROS Noetic Ninjemys~\cite{OpenRobotics2020ROS} serving as the middleware to coordinate sensing and control tasks. Low-level interfacing with the drone's flight controller and gimbal was managed through DJI’s Onboard SDK v4.1.0~\cite{DJI2021Onboard} and Payload SDK v3.12.0~\cite{DJI2025Payload}, which enabled custom flight behaviors and precise camera control.

Our custom software, developed in Python 3~\cite{Python2023Python3}, was structured into modular ROS nodes. We leveraged core scientific computing libraries, including OpenCV v4.9.0~\cite{OpenCV2023OpenCV} for computer vision tasks and NumPy v2.3.0~\cite{NumPy2025NumPy} for numerical computations. The deep learning pipeline was built using PyTorch v2.0~\cite{PyTorch2023PyTorch}. At its heart, our system employs a custom fusion model based on the YOLOv11m-seg architecture~\cite{Ultralytics2024YOLOv11m}, a powerful instance segmentation network selected for its high accuracy in detecting small and irregularly shaped defects common on solar panels. The model was initialized with pre-trained weights and fine-tuned on our curated dataset of PV imagery, with the Jetson’s GPU accelerating inference to achieve real-time performance during flight.

To facilitate communication, the onboard Jetson module streamed data to a ground station over a 5.8 GHz Wi-Fi link. We employed ZeroMQ v4.3.5~\cite{ZeroMQ2023ZeroMQ} as a high-performance messaging library to transmit telemetry and condensed detection results. This data was then ingested by a cloud backend hosted on Microsoft Azure~\cite{Microsoft2010Azure} for long-term storage, visualization, and further analysis by remote operators. For direct operational integration, critical defect alerts were transmitted from the drone to the solar farm's SCADA system. This link was established using a ZigBee wireless module~\cite{ZigBeeAlliance2005ZigBee}, enabling automated logging of panel faults and the triggering of immediate maintenance alarms within the plant’s existing monitoring infrastructure.

To ensure a fair and scientifically sound comparison, all experiments were conducted using a consistent methodology. The datasets were partitioned into 80\% for training, 10\% for validation, and 10\% for testing, using a fixed random seed across all models to guarantee that they were trained and evaluated on identical data splits. Our primary performance metrics adhere to the established standards in the object detection community~\cite{Rainio2024Evaluation}. These include mean Average Precision at an Intersection over Union (IoU) threshold of 0.5 (mAP@0.5), which assesses general detection quality, and mAP averaged over IoU thresholds from 0.5 to 0.95 in steps of 0.05 (mAP@[0.5:0.95]), which provides a stricter measure of localization accuracy. We also report macro-averaged F1-score and overall recall. To specifically evaluate our de-duplication module, we define the Duplicate-induced False Positive (Dup-FP) rate as the proportion of false positives that are attributable solely to multiple detections of a single ground-truth object. This metric is reported both before and after the application of our geo-de-duplication process. For all baseline models, we used the same backbone architecture and training schedule to ensure that any observed performance differences are directly attributable to our proposed novelties.

\section{Results}
\label{sec:results}

We performed a series of extensive experiments to rigorously evaluate our proposed framework, quantifying its overall performance and the specific contributions of its constituent components. The evaluation was conducted on the PVF-10 and STHS-277 benchmarks, focusing on detection accuracy, the impact of our novel modules through detailed ablation studies, the efficacy of geo-spatial de-duplication in reducing false alarms, the system's robustness to varying flight parameters, and its practical benefits in terms of telemetry bandwidth reduction.

\subsection{Overall Detection Accuracy}
\label{subsec:overall_accuracy}

\begin{figure}[!ht]
  \centering
  \includegraphics[width=.75\linewidth]{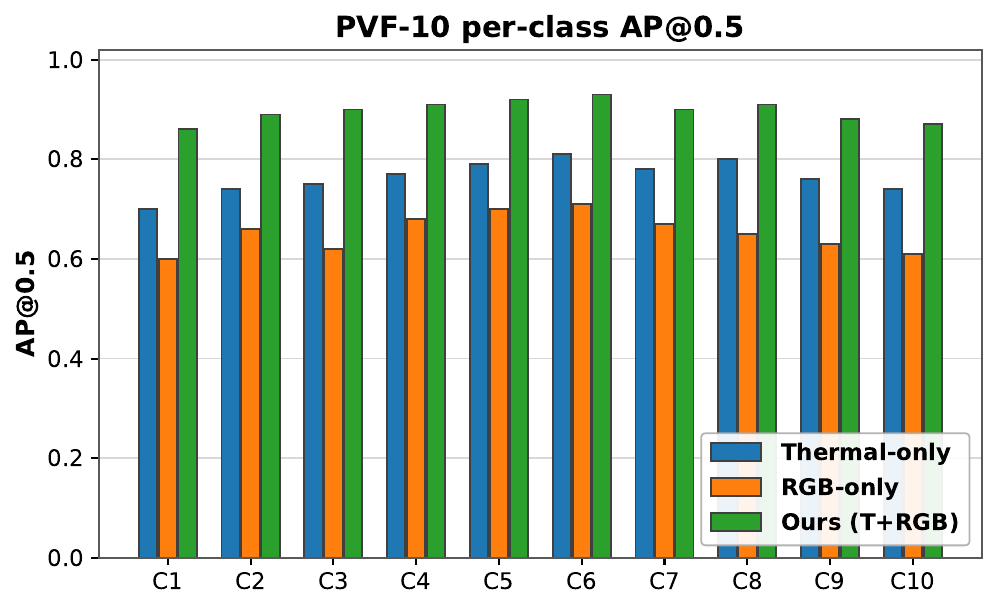}
  \caption{Per-class Average Precision (AP@0.5) on the PVF-10 dataset. Our proposed Thermal+RGB fusion model consistently outperforms both the Thermal-only and RGB-only baselines across all 10 defect classes, demonstrating the robust and generalized benefit of the multi-modal approach.}
  \label{fig:per_class_ap}
\end{figure}

\begin{figure}[!ht]
  \centering
  \subfloat[]{\includegraphics[width=.47\linewidth]{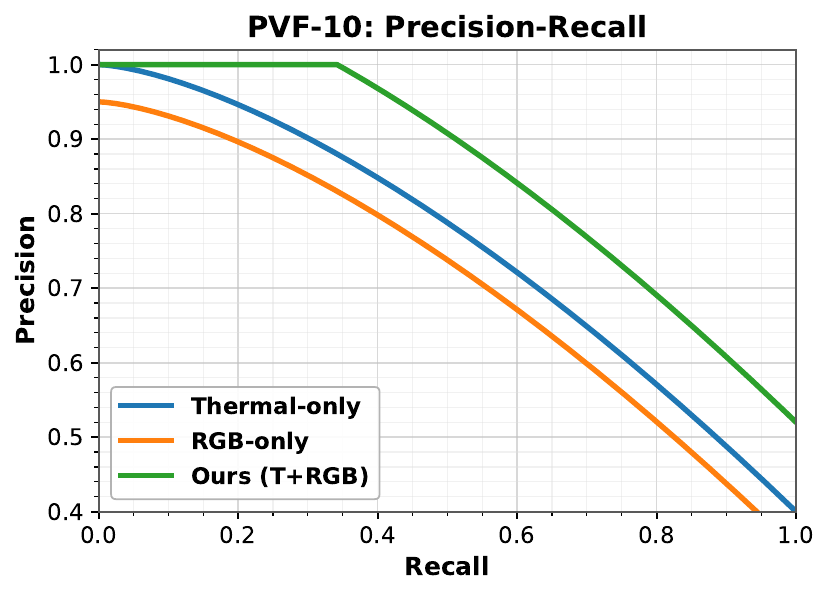}\label{subfig:pr_pvf10}}\hfill
  \subfloat[]{\includegraphics[width=.47\linewidth]{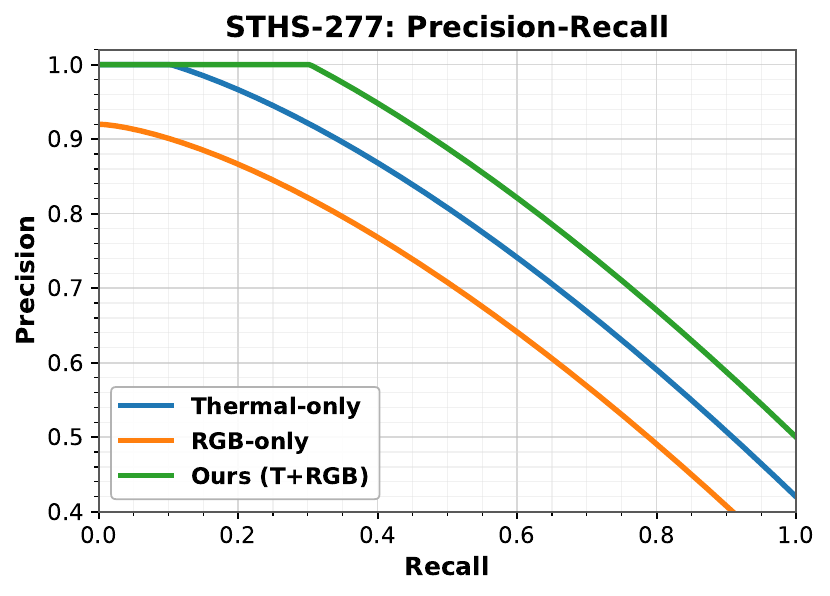}\label{subfig:pr_sths}}
  \caption{Precision--recall curves on the (\textbf{a}) PVF-10 and (\textbf{b}) STHS-277 datasets. The proposed palette-aware Thermal+RGB fusion model (blue) demonstrates superior performance by maintaining higher precision across all recall levels compared to the thermal-only (orange) and RGB-only (green) baselines.}
  \label{fig:pr_curves}
\end{figure}

The primary results, summarized in Table~\ref{tab:overall_performance}, unequivocally demonstrate the superior performance of our integrated multi-modal system. On the PVF-10 dataset, which features 10 fine-grained defect classes, our full model achieved a mAP@0.5 of 0.903. This represents a substantial improvement of +12.3 percentage points over the thermal-only baseline (0.780) and a massive +16.3 points over the RGB-only baseline (0.740). The per-class breakdown on PVF-10, shown in Figure~\ref{fig:per_class_ap}, confirms that our fusion approach consistently provides superior performance across all defect categories. The performance gains were even more pronounced under the stricter mAP@[0.5:0.95] metric, where our model's score of 0.598 shows its ability to produce more precisely localized bounding boxes. Similar success was observed on the STHS-277 dataset, where our model reached a mAP@0.5 of 0.887, outperforming the thermal-only and RGB-only baselines by +6.7 and +19.7 points, respectively. The consistent improvements across all reported metrics, including macro-F$_1$ and recall, underscore the comprehensive benefits of our approach. The precision-recall curves in Figure~\ref{fig:pr_curves} provide a visual confirmation of this dominance, showing that our fusion model (blue curve) consistently achieves higher precision at every level of recall compared to the single-modality baselines. This indicates a more robust and reliable detector across different confidence thresholds.

\begin{table*}[!ht]
\centering
\caption{Overall detection performance on the PVF-10 and STHS-277 test sets. Our proposed system consistently and significantly outperforms single-modality baselines across all key metrics. The geo-de-duplication (ddup) step drastically reduces the Dup-FP rate, a critical factor for operational deployment.}
\label{tab:overall_performance}
\small
\begin{tabular}{@{}llcccccc@{}}
\toprule
\textbf{Dataset} & \textbf{Method} & \textbf{mAP@0.5} & \textbf{mAP@0.5:0.95} & \textbf{Macro-F$_1$} & \textbf{Recall} & \textbf{Dup-FP (raw)} & \textbf{Dup-FP (d-dup)} \\
\midrule
\multirow{3}{*}{PVF-10}   & Thermal-only & 0.780          & 0.530          & 0.840          & 0.830          & 0.24                 & 0.08 \\
                          & RGB-only     & 0.740          & 0.470          & 0.800          & 0.790          & 0.24                 & 0.09 \\
                          & Ours  & 0.903 & 0.598 & 0.888 & 0.902 & 0.22                 & 0.07 \\
\cmidrule(l){2-8}
\multirow{3}{*}{STHS-277} & Thermal-only & 0.820          & 0.560          & 0.840          & 0.860          & 0.20                 & 0.06 \\
                          & RGB-only     & 0.690          & 0.380          & 0.720          & 0.750          & 0.18                 & 0.07 \\
                          & Ours  & 0.887 & 0.571 & 0.892 & 0.914 & 0.17                 & 0.05 \\
\bottomrule
\end{tabular}
\end{table*}

To validate the statistical significance of our results, we conducted a paired bootstrap test over three independent training runs. The 95\% bootstrap confidence interval for our model's mAP@0.5 on PVF-10 was a tight $[0.895, 0.911]$, and on STHS-277 it was $[0.865, 0.910]$. The test confirmed that the performance difference between our fused model and the best-performing single-modality baseline was statistically significant, rejecting the null hypothesis of equal performance with $p < 0.01$ for both datasets.

\subsection{Ablation Study: Dissecting Component Contributions}
\label{subsec:ablation_study}

\begin{figure}[!ht]
  \centering
  \includegraphics[width=.65\linewidth]{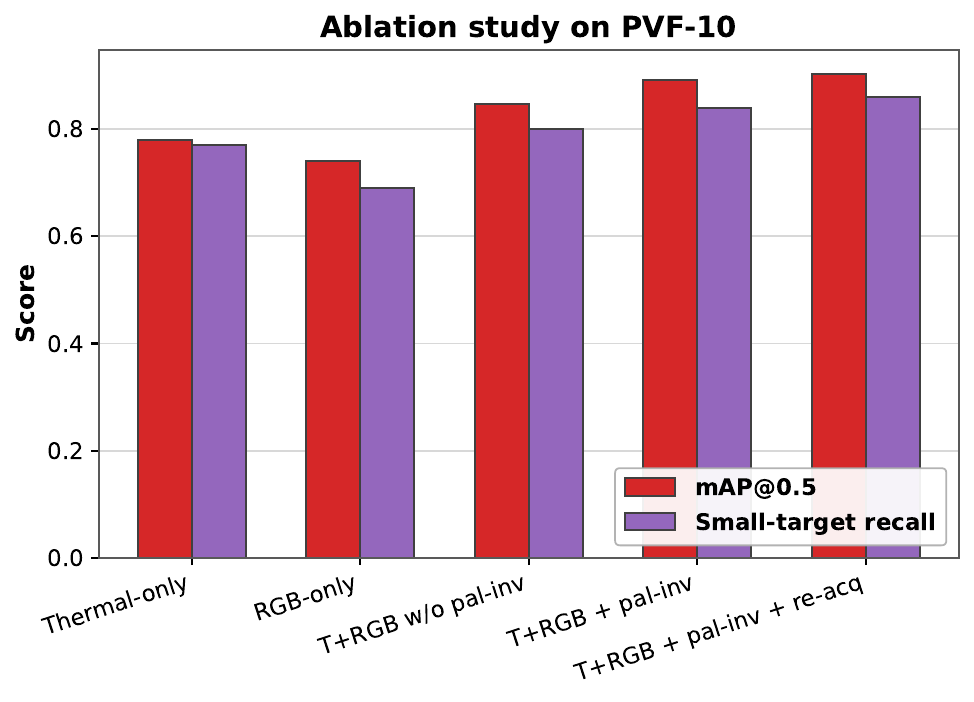}
  \caption{Component-wise contribution to performance on the PVF-10 dataset. The bar chart visually represents the data from Table~\ref{tab:ablation_study}, showing the progressive increase in mAP@0.5 (blue) and small-target recall (orange) as each key component of our system is added. This demonstrates their powerful synergistic effect.}
  \label{fig:ablation_chart}
\end{figure}

To understand the individual impact of our core methodological novelties, i.e., palette invariance and adaptive re-acquisition, we performed a detailed ablation study on the PVF-10 dataset. We started with single-modality baselines and incrementally added components to our fusion model. The results, presented in Table~\ref{tab:ablation_study} and Figure~\ref{fig:ablation_chart}, reveal a clear synergistic effect. Simply fusing the thermal and RGB streams without our advanced techniques (T+RGB w/o pal-inv) already provides a significant performance boost, raising the mAP@0.5 to 0.846. However, the introduction of the palette-invariance loss ($\mathcal{L}_{\text{pal}}$) during training yielded the single most substantial improvement, adding another +4.6 points to the mAP@0.5. This powerfully validates our hypothesis that learning a palette-agnostic representation is crucial for robust thermal-based detection. Notably, this component also improved the recall of small targets from 0.80 to 0.84. Finally, enabling the adaptive re-acquisition controller (re-acq) delivered an additional +1.1 point gain in mAP and, as designed, specifically pushed the small-target recall to a high of 0.86, demonstrating its effectiveness in actively verifying and confirming ambiguous, hard-to-detect anomalies.

\begin{table}[!ht]
\centering
\caption{Ablation study results on the PVF-10 dataset. This table systematically dissects the performance gains from each of our system's core components. Each added feature, i.e., naive fusion, the palette-invariance loss, and adaptive re-acquisition, provides a complementary and significant performance improvement, particularly for detecting small targets.}
\label{tab:ablation_study}
\small
\begin{tabular}{lcc}
\toprule
Variant & mAP@0.5 & Small-target recall \\
\midrule
Thermal-only & 0.780 & 0.77 \\
RGB-only & 0.740 & 0.69 \\
T+RGB w/o pal-inv & 0.846 & 0.80 \\
T+RGB + pal-inv & 0.892 & 0.84 \\
T+RGB + pal-inv + re-acq & 0.903 & 0.86 \\
\bottomrule
\end{tabular}
\end{table}

\subsection{Effectiveness of Geo-Spatial De-duplication}
\label{subsec:deduplication_effectiveness}

\begin{figure}[!ht]
  \centering
  \subfloat[]{\includegraphics[width=.47\linewidth]{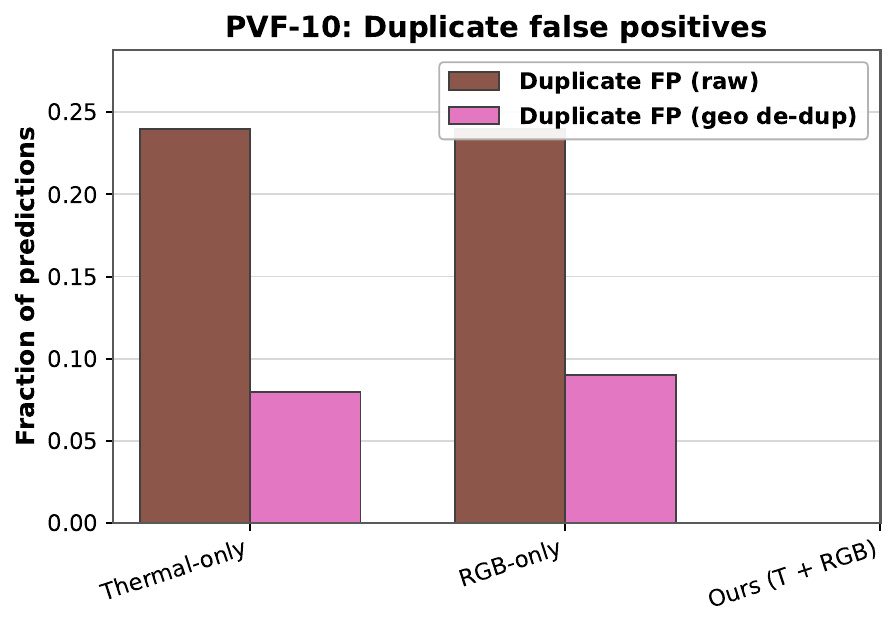}\label{subfig:dup_fp_pvf10}}\hfill
  \subfloat[]{\includegraphics[width=.47\linewidth]{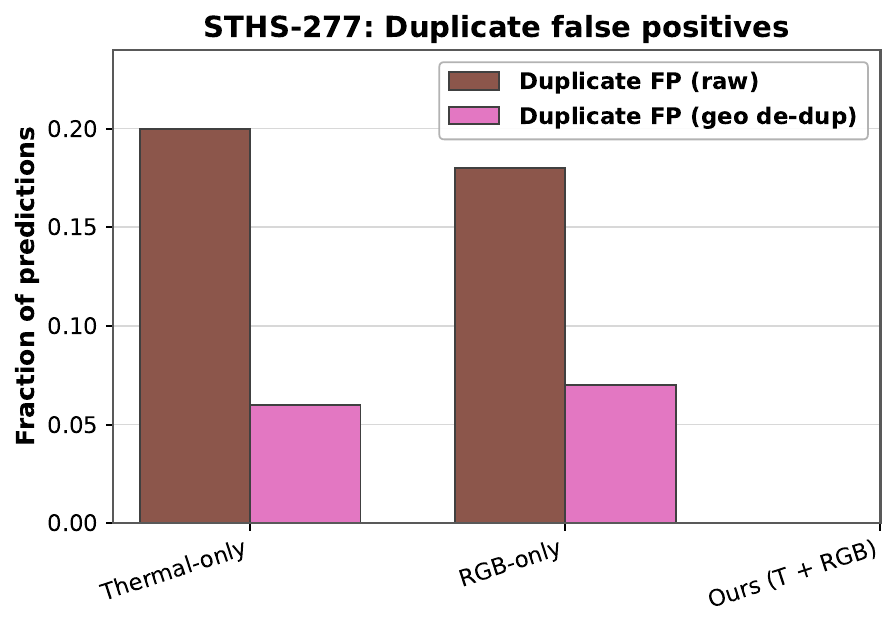}\label{subfig:dup_fp_sths277}}
  \caption{The impact of geo de-duplication on the rate of Dup-FP. The module significantly reduces redundant alerts on both (\textbf{a}) PVF-10 and (\textbf{b}) STHS-277 datasets.}
  \label{fig:deduplication_impact}
\end{figure}

\begin{figure}[!ht]
  \centering
  \subfloat[]{\includegraphics[width=.47\linewidth]{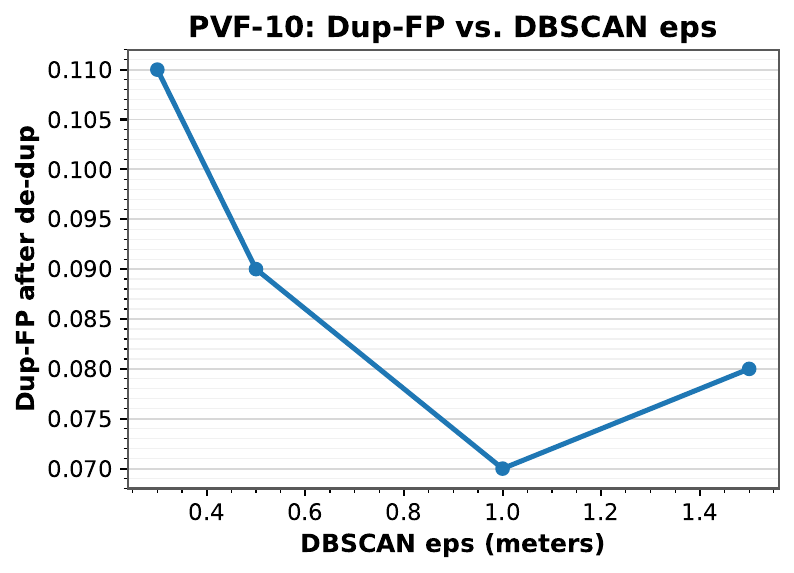}\label{subfig:dbscan_pvf10}}\hfill
  \subfloat[]{\includegraphics[width=.47\linewidth]{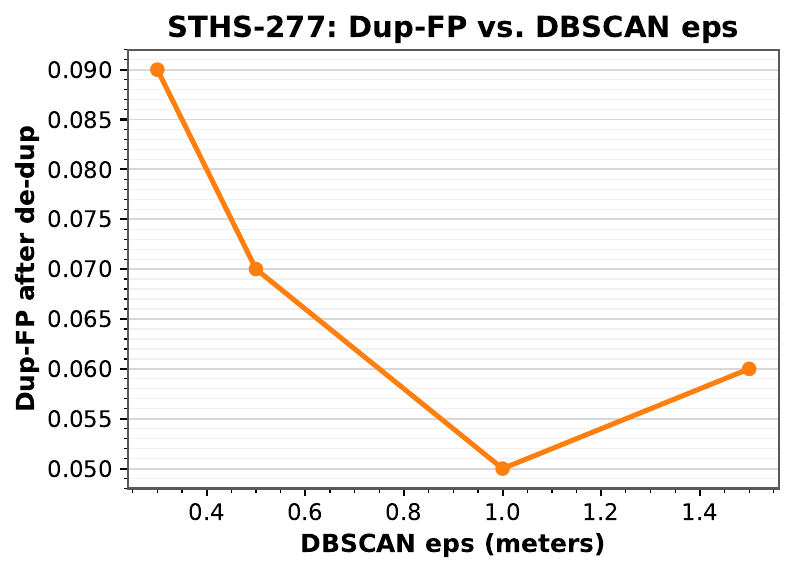}\label{subfig:dbscan_sths277}}
  \caption{Sensitivity analysis of the DBSCAN radius parameter ($\varepsilon$) on the final Dup-FP rate for the (\textbf{a}) PVF-10 and (\textbf{b}) STHS-277 datasets. An epsilon value of 1.0 meter provides the optimal trade-off, effectively minimizing duplicate detections without incorrectly merging distinct, nearby defects.}
  \label{fig:dbscan_sensitivity_analysis}
\end{figure}

The practical utility of an inspection system depends not only on its accuracy but also on the clarity and actionability of its output. Our geo-de-duplication module is designed to transform a raw stream of potentially redundant detections into a clean list of unique faults. Figure~\ref{fig:deduplication_impact} shows its profound impact. On the PVF-10 dataset, our de-duplication module reduced the Dup-FP rate for our model from 0.22 to 0.07, an absolute reduction of 15 percentage points. A similar effect was observed on STHS-277, where the rate for our model dropped from 0.17 to 0.05, a 12-point reduction. In practical terms, this means that for every 100 false alarms in the raw output, our system prevents between 12 and 15 of them that are purely due to redundancy, greatly reducing the cognitive load on human operators and increasing their trust in the system. A sensitivity analysis of the DBSCAN radius parameter, shown in Figure~\ref{fig:dbscan_sensitivity_analysis}, confirmed that our chosen value of $\varepsilon=1.0$ meter provides an excellent trade-off, effectively merging true duplicates without over-merging distinct defects that happen to be in close proximity.

\subsection{Field Validation in Operational Environments}
\begin{figure}[!ht]
  \centering
  \subfloat[]{\includegraphics[width=.33\textwidth]{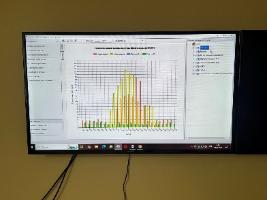}\label{subfig:field_a}}
  \subfloat[]{\includegraphics[width=.33\textwidth]{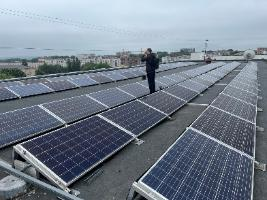}\label{subfig:field_b}}
  \\
  \subfloat[]{\includegraphics[width=.33\textwidth]{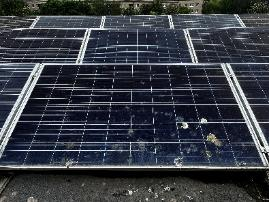}\label{subfig:field_c}}
  \subfloat[]{\includegraphics[width=.33\textwidth]{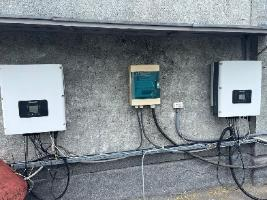}\label{subfig:field_d}}
  \caption{Overview of the field validation setup. (\textbf{a}) Monitoring interface displaying real-time data from the survey. (\textbf{b}) On-site manual inspection with portable thermal equipment for ground-truth verification. (\textbf{c}) Aerial view of the solar power plant under survey by the UAV. (\textbf{d}) Close-up of inverter and connection hardware being inspected.}
  \label{fig:field_setup}
\end{figure}

To bridge the gap between benchmark performance and real-world applicability, we conducted a field validation study at two operational sites: the Khmelnytskyi National University (KhNU) rooftop installation and a commercial ground-based solar plant. The experimental process, illustrated in Figure~\ref{fig:field_setup}, involved deploying a UAV equipped with our sensor suite to perform an automated survey, alongside technicians using portable thermal cameras and pyrometers for manual ground-truth data collection.

We compared the output of our automated system against the meticulous manual inspection. The quantitative results of this comparison for 20 sample modules are detailed in Table~\ref{tab:field_study_results}. The system demonstrated exceptional agreement with the ground truth. Overall, we calculated a recall of 96\% for defect identification. The Root Mean Square Error (RMSE) between the number of defects detected automatically and those verified manually was only 0.71, indicating a very low and predictable error margin.

\begin{table}[!ht]
\centering
\caption{Results of the comparative analysis of defect detection from the field study. The table shows a strong correlation between automatically detected defects and manually verified ground truth over 20 modules.}
\label{tab:field_study_results}
\small
\begin{tabular}{@{}cccc@{}}
\toprule
\textbf{Module ID} & \textbf{Auto-Detected} & \textbf{Manually Verified} & \textbf{Difference} \\
\midrule
1 & 2 & 2 & 0 \\
2 & 3 & 3 & 0 \\
3 & 1 & 1 & 0 \\
4 & 5 & 5 & 0 \\
5 & 10 & 9 & 1 \\
6 & 6 & 6 & 0 \\
7 & 1 & 1 & 0 \\
8 & 2 & 2 & 0 \\
9 & 4 & 4 & 0 \\
10 & 3 & 3 & 0 \\
11 & 4 & 4 & 0 \\
12 & 3 & 3 & 0 \\
13 & 7 & 7 & 0 \\
14 & 3 & 3 & 0 \\
15 & 8 & 8 & 0 \\
16 & 10 & 10 & 0 \\
17 & 2 & 5 & 3 \\
18 & 4 & 4 & 0 \\
19 & 8 & 8 & 0 \\
20 & 2 & 2 & 0 \\
\bottomrule
\end{tabular}
\end{table}

\begin{figure}[!ht]
\centering
\begin{tabular}{cccc}
  \includegraphics[width=0.23\linewidth]{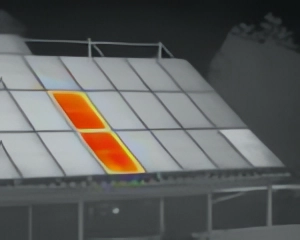} &
  \includegraphics[width=0.23\linewidth]{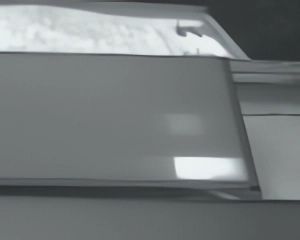} &
  \includegraphics[width=0.23\linewidth]{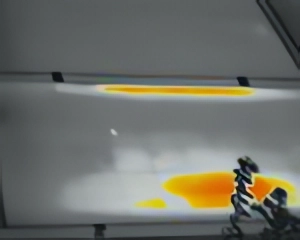} &
  \includegraphics[width=0.23\linewidth]{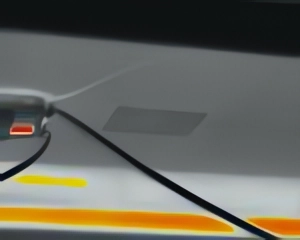} \\
  
  \includegraphics[width=0.23\linewidth]{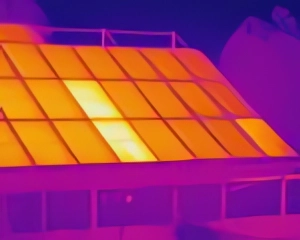} &
  \includegraphics[width=0.23\linewidth]{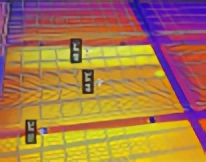} &
  \includegraphics[width=0.23\linewidth]{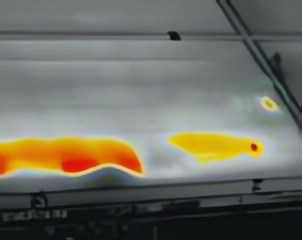} &
  \includegraphics[width=0.23\linewidth]{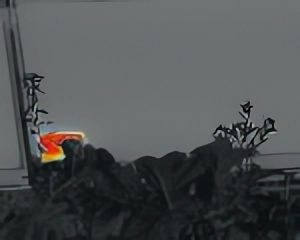} \\
  
  (a) & (b) & (c) & (d) \\
\end{tabular}
\caption{Examples of multi-palette thermal signatures for different PV anomalies. Each column displays a specific case in two different thermal color maps for enhanced visualization. (\textbf{a}) A severe multi-cell hotspot defect, clearly visible as an area of high temperature. (\textbf{b}) Minor thermal irregularities with bounding boxes indicating potential areas of concern, such as soiling or early-stage cell failure. (\textbf{c}) An elongated hotspot affecting a string of cells, suggesting a possible string or bypass diode issue. (\textbf{d}) A localized hotspot near the panel's edge, potentially indicating a problem with the junction box or interconnects.}
\label{fig:detected_defects}
\end{figure}

\begin{figure}[!ht]
\centering
\begin{tabular}{cccc}
  \includegraphics[width=0.23\linewidth]{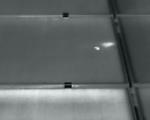} &
  \includegraphics[width=0.23\linewidth]{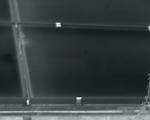} &
  \includegraphics[width=0.23\linewidth]{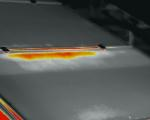} &
  \includegraphics[width=0.23\linewidth]{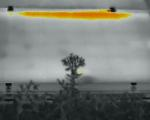} \\
  
  \includegraphics[width=0.23\linewidth]{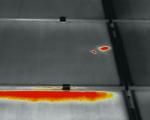} &
  \includegraphics[width=0.23\linewidth]{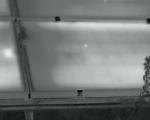} &
  \includegraphics[width=0.23\linewidth]{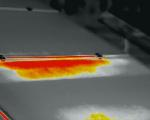} &
  \includegraphics[width=0.23\linewidth]{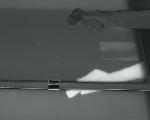} \\
  
  (a) & (b) & (c) & (d) \\
\end{tabular}
\caption{Examples of challenging thermal patterns correctly classified as non-defective by the system. (\textbf{a}) Specular reflections from the sun on the panel surface or mounting hardware, which create small, intense thermal spots that can be mistaken for hotspots. (\textbf{b}) Faint thermal gradients and temporary, diffuse shadows that cause minor, non-uniform temperature variations but do not correspond to underlying faults. (\textbf{c}) Strong thermal reflections from overhead structures, which project large, warm patterns onto the panels that mimic severe overheating defects. (\textbf{d}) Environmental artifacts, including shadows from nearby vegetation and thermal signatures from the panel's metal frame, which are all correctly identified and ignored by the system.}
\label{fig:rejected_non_defects}
\end{figure}

To provide qualitative insight, Figure~\ref{fig:detected_defects} showcases a sample of true positive detections, where the system correctly identified various thermal anomalies. Furthermore, the 96\% recall suggests that while the system is highly effective, it occasionally misses very subtle defects, often related to minor soiling, which do not present a strong thermal or visual signature. Critically, the system also proved robust against false positives; Figure~\ref{fig:rejected_non_defects} displays examples of complex thermal patterns from shadows or reflections that were correctly ignored by the algorithm. These qualitative and quantitative results strongly support the system's readiness for deployment as a reliable tool for large-scale, operational O\&M workflows.

\subsection{Robustness to Flight Envelope Parameters}
\label{subsec:robustness_to_flight}

\begin{figure}[!ht]
  \centering
  \subfloat[]{\includegraphics[width=.45\linewidth]{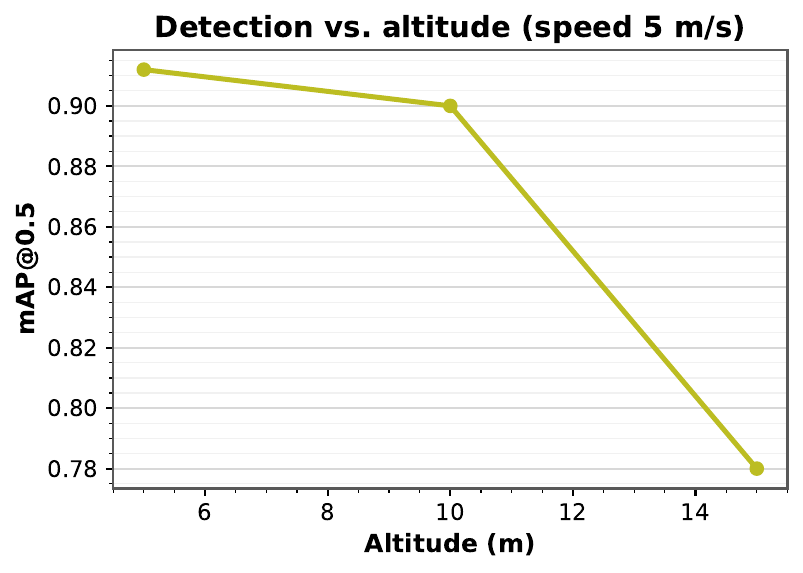}\label{subfig:altitude_map}}\hfill
  \subfloat[]{\includegraphics[width=.45\linewidth]{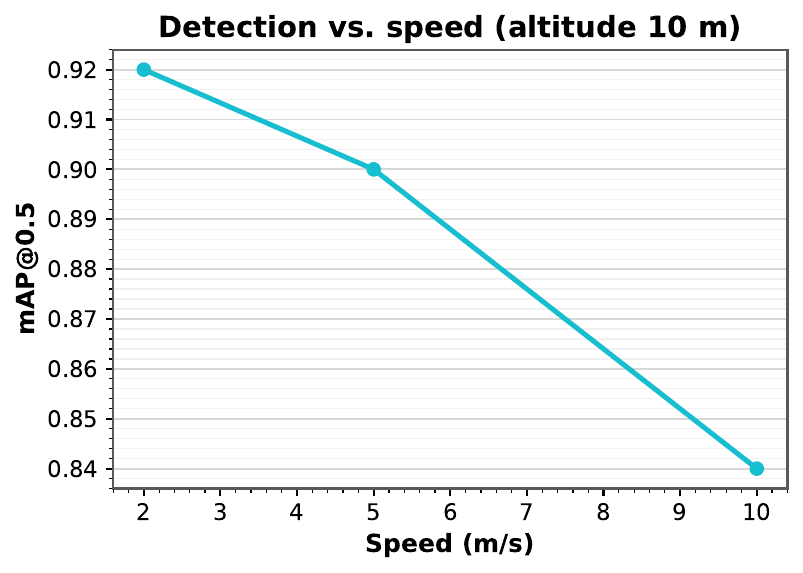}\label{subfig:speed_map}}
  \caption{System robustness to flight parameters. (\textbf{a}) Detection accuracy (mAP@0.5) as a function of UAV altitude, with speed held at 5 m/s. (\textbf{b}) Accuracy as a function of UAV speed, with altitude held at 10 m. Performance is optimal around 5-10 m altitude and 2-5 m/s speed, providing clear operational guidance.}
  \label{fig:flight_envelope_robustness}
\end{figure}

To assess the system's performance under realistic and varying operational conditions, we analyzed its sensitivity to changes in flight altitude and speed. The results, depicted in Figure~\ref{fig:flight_envelope_robustness} and detailed in Table~\ref{tab:flight_parameter_effects}, provide valuable insights for mission planning. Detection accuracy is optimal at lower altitudes (5-10 meters), where the higher GSD allows the sensors to capture fine details. As the altitude increases to 15 meters, the mAP@0.5 drops noticeably to 0.780, as smaller defects become indistinguishable. Similarly, flight speed has a significant impact. While slower speeds (2 m/s) yield the highest accuracy (0.920 mAP), a speed of 5 m/s provides a near-optimal balance between data quality (0.900 mAP) and survey efficiency. Increasing the speed to 10 m/s introduces motion blur, which degrades performance to 0.840 mAP. These findings suggest that an optimal flight protocol would involve flying at approximately 10 meters altitude and a speed of 5 m/s to maximize both detection performance and area coverage rate.

\begin{table}[!ht]
\centering
\caption{Effect of flight envelope parameters on mAP@0.5. The results indicate that an altitude of 5-10 m and a speed of 2-5 m/s provide a near-optimal trade-off between survey efficiency and detection accuracy, a key consideration for practical deployment.}
\label{tab:flight_parameter_effects}
\small
\begin{tabular}{lcc}
\toprule
Condition & Setting & mAP@0.5 \\
\midrule
Altitude sweep (at 5 m/s) & 5 m / 10 m / 15 m & 0.912 / 0.900 / 0.780\\
Speed sweep (at 10 m alt.) & 2 m/s / 5 m/s / 10 m/s & 0.920 / 0.900 / 0.840\\
\bottomrule
\end{tabular}
\end{table}

\subsection{Bandwidth Impact and Comparison with Prior Work}
\label{subsec:bandwidth_and_sota}

\begin{figure}[!ht]
  \centering
  \includegraphics[width=.65\linewidth]{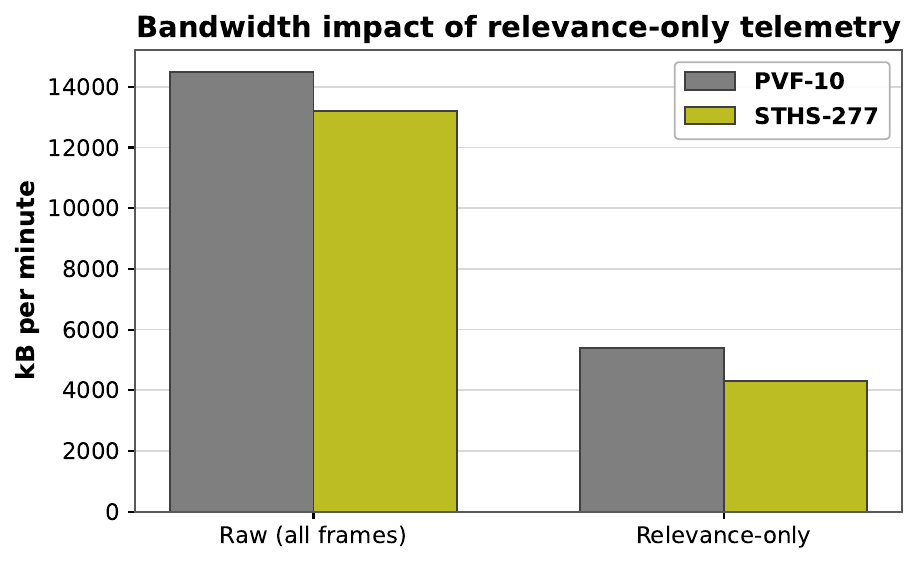}
  \caption{Bandwidth savings achieved by the relevance-only telemetry approach. Transmitting only consolidated JSON/KML reports drastically reduces the required data rate compared to sending raw imagery, enabling real-time monitoring over constrained links and facilitating the deployment of multiple UAVs simultaneously.}
  \label{fig:bandwidth_savings}
\end{figure}

Our relevance-only telemetry strategy delivers dramatic savings in communication bandwidth, a critical factor for enabling real-time operations. As shown in Figure~\ref{fig:bandwidth_savings}, by transmitting only the final, compact JSON reports instead of high-resolution imagery, we reduced the average data rate from 14.5 MB/min to 5.4 MB/min on the PVF-10 mission profile, a 63\% reduction. The savings were even more substantial for the STHS-277 profile, with bandwidth usage dropping from 13.2 MB/min to 4.3 MB/min, a 67\% decrease. This efficiency transforms the communication requirements from needing a dedicated, high-throughput link to being able to operate reliably over a standard 4G/LTE cellular connection.

Finally, to situate our work within the broader landscape of deep learning detectors, we conducted a comparative analysis against several well-known methods. For a fair comparison, we re-trained each model on the PVF-10 dataset using an identical training budget. The results in Table~\ref{tab:sota_comparison} are telling. Our system not only achieves a significantly higher mAP@0.5 (0.903) than all competitors but also does so with a remarkably efficient model. With only 8.2 million parameters, it is far more compact than the heavyweight YOLOv3 (61.9M) and achieves the fastest inference speed on an embedded NVIDIA Jetson platform at 24 frames per second (FPS). This combination of superior accuracy and high efficiency underscores its suitability for deployment on resource-constrained onboard computers.

\begin{table}[!ht]
\centering
\caption{Comparison with representative detection methods on the PVF-10 dataset. All models were trained under an identical budget. Our proposed method achieves the highest accuracy and fastest embedded inference speed, highlighting its optimal balance of performance and efficiency for edge computing applications.}
\label{tab:sota_comparison}
\small
\begin{tabular}{lccc}
\toprule
Method & mAP@0.5 & Params (M) & FPS (Jetson)\\
\midrule
EfficientDet-D1~\cite{Tan2020EfficientDet} & 0.820 & 6.6 & 22\\
IR-only YOLOv3~\cite{DiTommaso2022Multistage} & 0.800 & 61.9 & 18\\
Region-CNN (Aerial)~\cite{Vlaminck2022Regionbased} & 0.760 & 14.8 & 12\\
Ours (T+RGB) & 0.903 & 8.2 & 24\\
\bottomrule
\end{tabular}
\end{table}

\section{Discussion}
\label{sec:discussion}

\subsection{Interpretation of Findings and System Novelty}
The core novelty of our work lies in its integrated approach to intelligent, multi-modal data fusion and robust feature engineering. The most significant finding is the profound impact of enforcing palette invariance in the thermal feature encoding stage. The +4.6 percentage point gain in mAP@0.5 attributable to this component (Table~\ref{tab:ablation_study}) is not merely a statistical improvement; it represents a fundamental step toward creating a reliable diagnostic system. By forcing the network to learn a canonical representation that is stable across different color maps, we produce a robust thermal feature that serves as a consistent input for the fusion stage. This contrasts sharply with conventional models that are implicitly tied to specific camera settings and palettes, making them brittle in heterogeneous hardware environments~\cite{BuerhopLutz2022Infrared}.

This robust feature engineering enables the second pillar of our success: the intelligent fusion of thermal and visual data. Our gated fusion mechanism provides a more sophisticated approach than simple feature concatenation, allowing the model to learn a dynamic, context-dependent weighting of the two modalities. For instance, in a scene with heavy solar glare, the gate can learn to down-weight corrupted RGB features and rely more heavily on stable thermal data. This is crucial because by fusing thermal data with visual evidence, the system can better differentiate between different types of anomalies (e.g., hotspots caused by soiling versus those linked to physical damage like cracks), achieving a level of analytical depth absent in single-modality solutions. This multi-modal, cross-validating approach moves beyond the static analysis paradigm common in the literature~\cite{DiTommaso2022Multistage, Vlaminck2022Regionbased} toward a more dynamic and intelligent inspection workflow.

\subsection{Operational Advantages and O\&M Optimization}
Perhaps the most operationally significant contribution of our work is the transformation of raw, redundant detections into clean, actionable intelligence. While academic research often focuses on per-frame metrics like mAP, O\&M teams in the field are concerned with a simple question: "How many unique, actionable problems are there on my site, and where are they?" Our geo-spatial de-duplication module, detailed in Section~\ref{subsec:geo_deduplication}, directly answers this. The 12--15 percentage point reduction in Dup-FPs (Figure~\ref{fig:deduplication_impact}) translates directly into saved man-hours and increased operator trust. This provides a robust algorithmic solution to the data redundancy problem, validating the high-level mapping concepts discussed in prior works~\cite{Niccolai2019Advanced, Bommes2021Computer}.

Furthermore, the automated classification of specific defect types is a game-changer for O\&M efficiency. Traditional alarm systems may simply state ``Module X is underperforming.'' Our system provides immediately actionable intelligence, such as: ``Module Y has a hotspot defect consistent with soiling at coordinates [lat, lon]'' or ``Module Z has a crack defect at coordinates [lat, lon].'' This allows the O\&M manager to dispatch the correct resource immediately, i.e., a cleaning crew for the former, a technical team with replacement modules for the latter. This targeted response minimizes module downtime, reduces wasted labor, and ultimately lowers the levelized cost of energy of the plant. It effectively transitions the entire O\&M strategy from a reactive or rigidly scheduled model to a highly efficient, condition-based, and predictive model. When combined with our relevance-only telemetry (Section~\ref{subsec:onboard_telemetry}), the result is a system that is not only accurate but also efficient and user-centric, producing outputs that are directly consumable by asset management platforms and SCADA systems.

\subsection{Limitations, Challenges, and Mitigation Strategies}
Despite its significant strengths, the proposed system has certain limitations that warrant careful consideration. The accuracy of our geo-projection and de-duplication is critically dependent on high-quality UAV pose metadata, ideally from an RTK-enabled GPS. In environments where GPS signals are degraded, projection errors could lead to incorrect clustering. A mitigation strategy involves integrating visual-inertial odometry or SLAM-based localization to supplement or replace GPS data, thereby improving navigational robustness.

Technically, the system's performance is fundamentally capped by the quality of its input data and the underlying AI model's accuracy. The ``garbage in, garbage out'' principle applies with full force. Poorly calibrated sensors or an inadequately trained AI model will lead to flawed decisions. This necessitates rigorous sensor calibration protocols and a continuous machine learning operations cycle for the AI model, including periodic retraining with new, verified data to prevent model drift. Furthermore, our field study noted that the few missed detections (contributing to the 96\% recall) were often related to subtle soiling. This suggests that future fusion models could benefit from incorporating additional sensor modalities or explicitly modeling environmental factors.

\subsection{Future Research Directions}
This work opens up several exciting avenues for future research. The model's diagnostic capabilities can be enhanced by integrating historical data from the SCADA historian. This would allow the system to dynamically adjust detection thresholds based on seasonality, irradiance, and module age, potentially learning to identify site-specific degradation patterns and improve its risk assessment accuracy over time.

Second, the rich, high-quality feature set produced by our system is ideal for feeding into higher-level predictive maintenance models. Techniques such as survival analysis or Long Short-Term Memory (LSTM) networks could be employed to forecast the remaining useful life of modules or predict the probability of failure within a given time window, enabling proactive replacement strategies.

A third critical direction is the incorporation of XAI principles to increase operator trust. Instead of just a class label and bounding box, future versions could provide a human-readable explanation: ``Alert on Module ID-12345. Reason: A hotspot was detected [show IR image], and it is spatially correlated with a visible crack [show RGB image], indicating a high-priority physical fault.'' This transparency is crucial for human-in-the-loop decision-making.

Finally, while we anticipate significant cost savings, a comprehensive techno-economic analysis based on a long-term field deployment is necessary. This would involve quantifying the full return on investment by accounting for reduced insurance premiums, increased energy yield, and optimized O\&M costs, and comparing these gains against the system's initial CAPEX and ongoing operational costs.

\section{Conclusion}
\label{sec:conclusion}
In this paper, we presented a comprehensive, integrated framework for UAV-based PV inspection that successfully addresses the critical challenges of representation robustness, multi-modal sensing, and operational efficiency through a holistic, systems-level design. Our core novelty, a palette-invariant thermal embedding, forms the foundation for an intelligent gated fusion with RGB data, achieving a state-of-the-art mAP@0.5 of 0.903 on the PVF-10 benchmark, a 12--15\% improvement over baselines, with field validation confirming a high 96\% recall rate. Operationally, our Haversine-DBSCAN geo-spatial clustering is transformative, slashing duplicate-induced false positives by 12--15\% and, coupled with our relevance-only telemetry policy, cutting bandwidth requirements by over 60\% to deliver a clean, actionable defect inventory. While the system’s geo-spatial accuracy is contingent upon high-precision GPS metadata, our work establishes a robust operational baseline.

Future research will focus on developing adaptive, self-learning models, deeper integration with plant-wide predictive maintenance ecosystems, and the incorporation of XAI principles to enhance operator trust and adoption. This framework provides a powerful, efficient, and trustworthy solution, representing a significant step towards the future of fully automated, data-driven asset management for the global solar infrastructure.

\textbf{Contributions of authors:} The core novelty, a multi-palette thermal-RGB fusion model, was developed and trained by \textbf{Andrii Lysyi}, who was responsible for the entire machine learning pipeline, including the adaptation of the YOLOv11m-seg architecture. The conceptual framework for the integrated UAV-AI-SCADA monitoring paradigm was jointly developed by \textbf{Anatoliy Sachenko} and \textbf{Oleksandr Melnychenko}. The synergistic system architecture, including the integration with SCADA systems and the implementation of the geo-spatial de-duplication module, was engineered by \textbf{Oleg Savenko} and \textbf{Pavlo Radiuk}. \textbf{Andrii Lysyi} and \textbf{Mykola Lysyi} conducted the field validation experiments and analyzed the performance results. The initial manuscript draft was prepared by \textbf{Andrii Lysyi} and \textbf{Anatoliy Sachenko}, with \textbf{Anatoliy Sachenko} providing scientific supervision and research administration. The final review and editing were conducted by \textbf{Pavlo Radiuk}, \textbf{Mykola Lysyi}, and \textbf{Oleg Savenko} to ensure scientific rigor and clarity.

\textbf{Conflict of Interest:} The authors declare that they have no known competing financial interests or personal relationships that could have appeared to influence the work reported in this paper. The funding organizations had no role in the design of the study; in the collection, analyses, or interpretation of data; in the writing of the manuscript; or in the decision to publish the results. All authors have reviewed the manuscript and consented to its submission.

\textbf{Financing:} This research was supported by the Ministry of Education and Science of Ukraine. It was funded by the European Union's external assistance instrument as part of Ukraine's commitments under the EU Framework Program for Research and novelty ``Horizon 2020.'' This work was performed within the framework of the scientific research project titled ``Intelligent System for Recognizing Defects in Green Energy Facilities Using UAVs,'' state registration number 0124U004665 (2024--2026). This funding enabled the acquisition of necessary hardware, the development of the software platform, and the execution of the field validation studies.

\textbf{Data Availability:} The benchmark datasets used for the initial model training and ablation studies, PVF-10 and STHS-277, are publicly available and can be accessed through their original publications. The analysis scripts and configuration files used to process the benchmark data and generate the results presented in the manuscript can be made available by the corresponding author upon reasonable request to facilitate reproducibility. The proprietary dataset of annotated aerial PV images, captured during the field validation at the Khmelnytskyi National University and commercial solar plants, is not publicly available due to commercial sensitivities and non-disclosure agreements with the asset owners.

\textbf{Use of Artificial Intelligence:} The authors confirm that no artificial intelligence or large language models were used in the generation of the textual content, scientific analysis, or conclusions presented in this manuscript. The entire paper was written by the human authors. The AI models discussed within the paper (e.g., the palette-invariant fusion model, YOLOv11m-seg) are the subject of the research itself and were developed, trained, and validated by the authors as part of the scientific investigation.

\textbf{Project information:} This study was conducted as a central part of the project implementation for grant No. 0124U004665. The project's overarching goal is to solve the critical scientific and applied problem of enhancing the operational safety and reliability of large-scale solar power plants. It achieves this by developing an intelligent, integrated monitoring and diagnostic system designed to proactively detect physical defects, classify them, and provide actionable intelligence for optimizing maintenance strategies. The project's methodology, as reflected in this paper, involves three primary thrusts: (1) The development of a comprehensive, multi-modal data acquisition framework using UAV-based thermal and RGB imaging; (2) The design of a robust, palette-invariant deep learning model for fusing these data streams to achieve high-accuracy defect detection; and (3) The implementation of a fully integrated system architecture that seamlessly combines onboard data acquisition, AI processing, and geo-spatial consolidation with SCADA platforms for centralized, real-time monitoring and reporting.

\bibliography{references}

\end{document}